\documentclass[letterpaper, 10 pt, conference]{ieeeconf}  %

\IEEEoverridecommandlockouts                              %

\title{\LARGE \bf
Learnable Spatio-Temporal Map Embeddings for Deep Inertial Localization
}

\author{Dennis Melamed, Karnik Ram, Vivek Roy, and Kris Kitani%
\thanks{All authors are with the Robotics Institute in Carnegie Mellon University, Pittsburgh, PA, USA.
        Correspondence: {\tt\small kkitani@cs.cmu.edu}}.\thanks{$^\dagger$Project page: \url{https://rebrand.ly/learned-map-prior}}}

\usepackage{subcaption}
\usepackage{graphicx}
\usepackage{amsmath}
\usepackage{booktabs}
\usepackage{cite}
\usepackage{multirow}
\usepackage{tabularx}
\usepackage{hyperref}
\usepackage{tikz}

\hypersetup{
    colorlinks=true,
    urlcolor=black}

\newcommand\copyrighttext{%
  \footnotesize \textcopyright 2022 IEEE. Personal use of this material is permitted.
  Permission from IEEE must be obtained for all other uses, in any current or future
  media, including reprinting/republishing this material for advertising or promotional
  purposes, creating new collective works, for resale or redistribution to servers or
  lists, or reuse of any copyrighted component of this work in other works.}
\newcommand\copyrightnotice{%
\begin{tikzpicture}[remember picture,overlay]
\node[anchor=south,yshift=10pt] at (current page.south) {\fbox{\parbox{\dimexpr\textwidth-\fboxsep-\fboxrule\relax}{\copyrighttext}}};
\end{tikzpicture}%
}

\begin{document}
\bstctlcite{Force_Etal}

\maketitle
\copyrightnotice
\thispagestyle{empty}
\pagestyle{empty}

\begin{abstract}

Indoor localization systems often fuse inertial odometry with map information via hand-defined methods to reduce odometry drift, but such methods are sensitive to noise and struggle to generalize across odometry sources.
To address the robustness problem in map utilization, we propose a data-driven prior on possible user locations in a map by combining learned spatial map embeddings and temporal odometry embeddings.
Our prior learns to encode which map regions are feasible locations for a user more accurately than previous hand-defined methods.
This prior leads to a 49\% improvement in inertial-only localization accuracy when used in a particle filter. This result is significant, as it shows that our relative positioning method can match the performance of absolute positioning using bluetooth beacons. To show the generalizability of our method, we also show similar improvements using wheel encoder odometry. Our code will be made publicly available$^\dagger$.

\end{abstract}

\section{Introduction}
Existing approaches to using occupancy map information to improve odometry-only localization are often hand-defined, leading to sensitivity to odometry errors and implicit assumptions which do not always generalize across odometry sources.
Methods like SLAM \cite{zhang_visual-lidar_2015}, map-matching \cite{rechy_rormero_map-aware_2018, xiao_lightweight_2014}, and heuristic map constraints on motion \cite{maclachlan_applied_nodate, xia_indoor_2019} show that maps are a rich information source to improve localization, but new methods are needed to more robustly utilize map information. We seek to extract more robust map information via a learned model to improve localization in odometry-only scenarios, particularly for indoor localization.

Inertial odometry utilizes inertial measurement units (IMUs) like those in smartphones (Fig. \ref{fig:zed_rig}) to localize agents.
The accuracy of inertial odometry has seen significant improvements from deep methods \cite{yan_ronin_2019,liu_tlio_2020, sun_idol_2021} but still suffers from drift due to integration of small errors in relative motion estimates.
This is unavoidable without introducing absolute constraints on user location, like those provided by maps.
Introducing map information allows for drift correction without requiring additional sensors to provide absolute constraints like cameras, LiDARs, or fingerprinting systems \cite{agarwal_deepble_2021}.

\begin{figure}[h]
    \centering
    \includegraphics[width=0.48\textwidth]{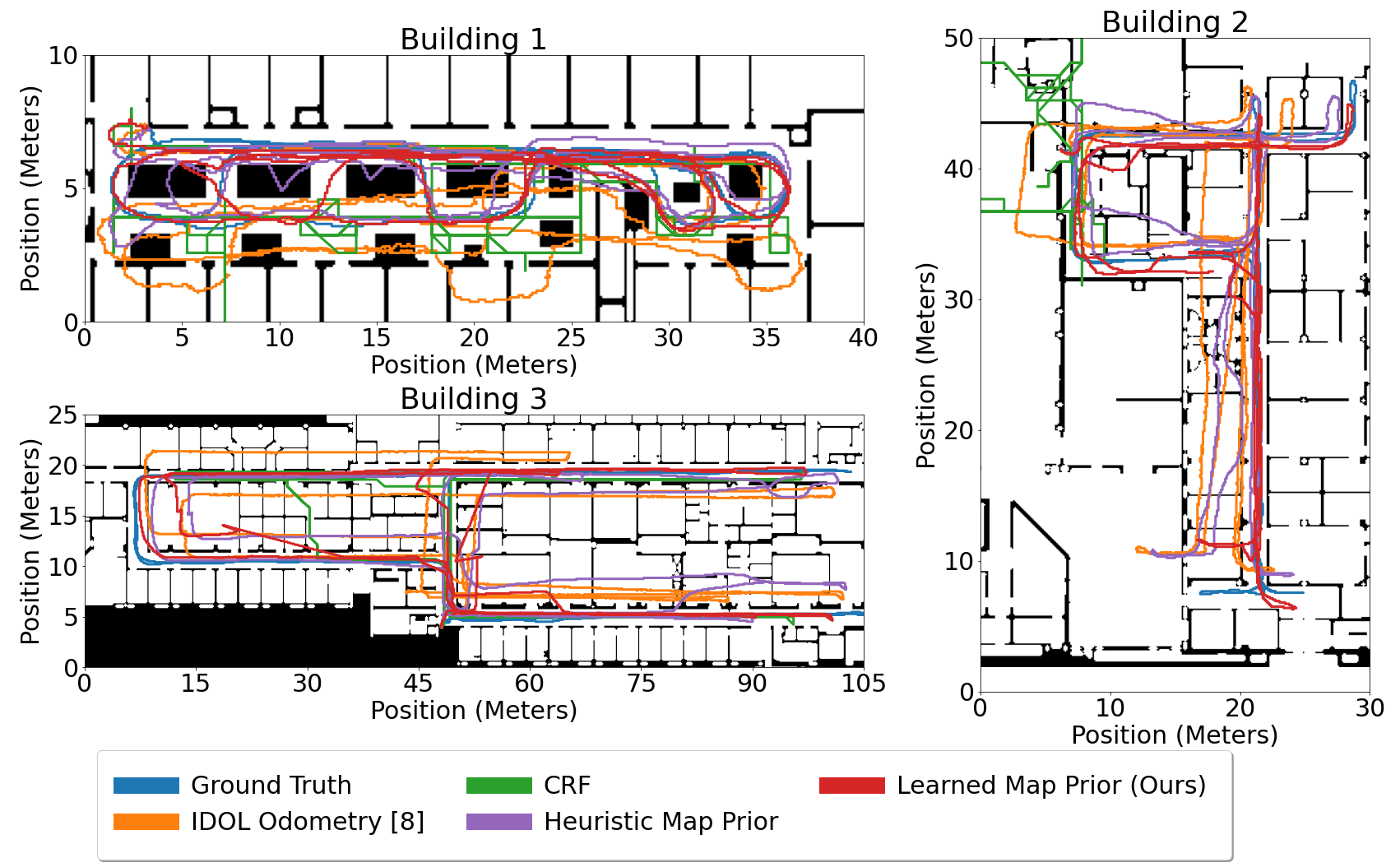}
    \caption{Our proposed method utilizes learnable spatio-temporal map priors to reduce drift in inertial odometry. Plots show qualitative comparison of trajectories estimated by proposed method against baselines on different buildings (IDOL dataset \cite{sun_idol_2021}).}
    \label{fig:IDOL_trajs}
    \vspace{-2em}
\end{figure}

Existing indoor localization systems using vision, LiDAR, and fingerprinting suffer from downsides that odometry-only methods avoid. Image or LiDAR-based localization via Simultaneous Localization and Mapping (SLAM) can be highly accurate \cite{zhang_visual-lidar_2015, zhang_loam_2014}, but can require well-lit and highly featured environments, be power-hungry, and be constraining for human activities since these sensors always have to be able to see the world. WiFi and Bluetooth fingerprinting systems  \cite{ahmetovic_navcog_2016, palaskar_wi-fi_2014} require instrumentation of environments, which can be costly and require significant upkeep. 
The deep inertial odometry methods proposed in RoNIN \cite{yan_ronin_2019}, TLIO \cite{liu_tlio_2020}, and IDOL \cite{sun_idol_2021} avoid the above drawbacks by making use of IMUs. IMUs are lightweight, use only milliAmps of current, do not require line-of-sight, and are present in most modern smartphones and robots. 

Map information has been utilized to reduce inertial odometry drift \cite{xia_indoor_2019,xiao_lightweight_2014, maclachlan_applied_nodate}, but prior methods are generally hand-defined. 
These hand-defined methods make assumptions which generalize poorly, like discretizations of the map for human localization which are too coarse for slower-moving robots, or use heuristics on location feasibility.
In our experimentation, these heuristics often lead to localization errors.
To reduce odometry drift without hand-defined methods, our approach fuses spatial map information and temporal odometry information via an efficient learnable map prior. Our deep network uses a convolutional model to encode map information and a recurrent model to encode recent odometry measurements. These encodings are combined into a map prior to determine areas of the map where the recent odometry measurements would have been feasible with higher accuracy than prior methods. When used as the sensor model of a simple particle filter, our prior is able to utilize highly noisy inertial odometry to provide approximately $49\%$ improvements in localization accuracy over the original odometry. This level of improvement is also seen when our system is applied to wheel encoder odometry on robots like in Fig. \ref{fig:tb_rig} without requiring any retraining.

\section{Related Work}

Indoor localization has been explored from several angles including pure odometry approaches, combining maps with odometry, and utilizing other sensors combined with odometry. 

\noindent\textbf{Pure Odometry Approaches:}
Inertial odometry has used both traditional and data-driven approaches. Traditional approaches generally use device heading estimation, step-detection, and heuristic stride length assumptions to perform Pedestrian Dead Reckoning (PDR) by adding up strides in the estimated user heading \cite{beauregard_indoor_2008}. IONET \cite{chen_ionet_2018} used a deep network to learn to consume IMU measurements and output user polar displacements. RoNIN \cite{yan_ronin_2019} explored ResNets, temporal convolutional networks, and recurrent networks to determine the best performing architectures for the inertial localization task, and TLIO \cite{liu_tlio_2020} added a Kalman filtering element to improve accuracy. 
IDOL \cite{sun_idol_2021} reduced inertial heading errors by using a learned deep orientation estimation system.
Wheel encoders are also a common source of pure odometry, and recent applications of deep networks have provided better uncertainty estimation to reduce encoder odometry drift \cite{onyekpe_whonet_nodate}. Despite improvements, most odometry-only approaches still suffer from drift over the long term.

\noindent\textbf{Map \& Odometry Fusion:}
Map information has been used for localization via several different techniques. Graph-based systems turn a map into a series of connected nodes which lie in the space the user might travel through. 
The user's trajectory is then constructed out of physically feasible transitions between these nodes which best match the user's odometry. 
Luo et al. \cite{luo_enhanced_2017} implement a hidden Markov model (HMM) to localize users' unstable GPS signals on a graph of outdoor road segments. Deep attention networks have been shown to outperform HMMs in some scenarios for matching user travel to road segments \cite{zhao_deepmm_2019}.
Indoors, conditional random fields (CRFs) have seen high pedestrian localization accuracy by placing nodes in physically accessible space, spaced apart by human step lengths \cite{xiao_lightweight_2014}.
Transitions in this graph are predicted based on inertial odometry. Graph approaches are successful, but generally require hand-design of the underlying graph and transition model for a specific source of odometry or sensor measurement. For example, \cite{xiao_lightweight_2014} implicitly assumes the desired localization precision is that of a human user's step by their node spacing. 
A small wheeled robot may need more granular localization, meaning that the map graph has to be completely regenerated to transfer to the new platform.

Bayesian filtering systems improve localization accuracy by combining information sources like odometry with absolute constraints like maps or absolute position measurements. 
Extended Kalman filtering (EKF) is an efficient approach to estimate agent state from both odometry and external information \cite{simon_optimal_2006}, but suffers from only maintaining one estimate of state and assumes Gaussian uncertainty. Particle filters \cite{liu_sequential_1998} are often used when state distributions are unknown, change over time, and could be multi-modal. 
Multiple estimates of location are propagated using odometry, then kept or removed based on information from maps or other sensors. 
Particle filters are simple and adaptable so lend themselves well to localization on low compute devices. 
For particle filters, map constraints are usually invoked by heuristics which remove particles whose last odometry propagation is infeasible according to the map (\emph{e.g.}, penetrating a wall).
Such a system has been proven useful in practical scenarios \cite{beauregard_indoor_2008, xia_indoor_2019, maclachlan_applied_nodate}, but removing unlikely particles in this way is extremely sensitive to certain types of odometry noise (\emph{e.g.,} proximity to obstacles can remove too many particles). 
The particle filter presented by \cite{rechy_rormero_map-aware_2018} computes overlap between map obstacles and a trajectory computed by integrating a few previous odometry measurements backwards from a particle location. 
Particles are given more weight if there are fewer overlaps. 
This type of multi-step heuristic is less sensitive to odometry noise than a single-step heuristic, but still suffers from the high levels of noise in inertial odometry.
Our work suggests that a deep network may able to capture more general shape characteristics of the odometry to better match map locations even when odometry error is high.

\noindent\textbf{Localization with other sensors:}
Vision and LiDAR systems like V-LOAM \cite{zhang_visual-lidar_2015} are highly accurate, but require line-of-sight to the environment and have significant power and computation needs. 
Bluetooth beacons or WiFi access points are also used to generate a pattern of radio frequency signals which can identify a location fairly precisely \cite{agarwal_deepble_2021, palaskar_wi-fi_2014}. Radio frequency systems are even more effective when combined with inertial odometry to smooth the noisy absolute location estimates \cite{ahmetovic_navcog_2016}. 
However, such systems often require expensive infrastructure installation and maintenance, and building materials can block signal propagation. Utilizing an odometry-only approach resolves many of these problems, particularly if the sensors are low-power and non-line-of-sight like the IMUs and wheel encoders used here.

\section{Method}
We aim to develop a system which learns to consume high drift odometry measurements and occupancy map information to reduce drift in the estimated user location.
Map information is integrated via the sensor model of a particle filter, but prior methods of doing so are sensitive to high levels of error in inertial odometry.
Instead, we learn a data-driven prior from odometry and maps which we hypothesize will indicate likely user locations with higher accuracy than prior heuristic methods.

\begin{figure}[h]
    \centering
    \includegraphics[width=\linewidth]{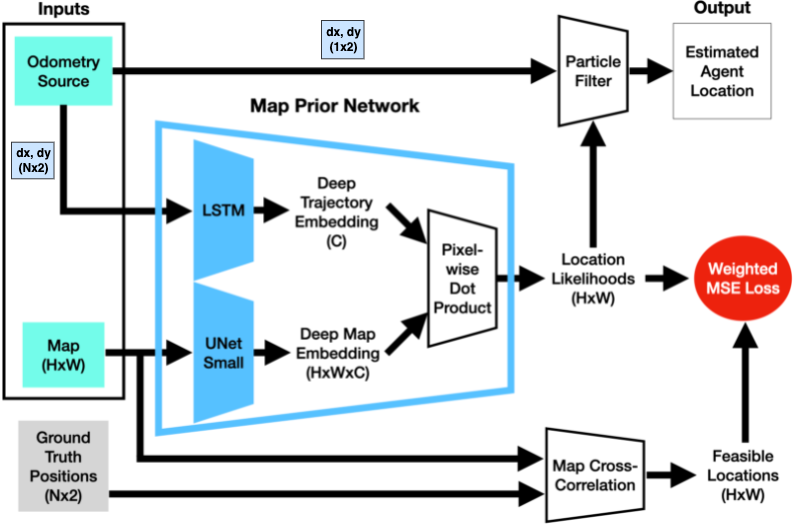}
       \caption{System Diagram: Our map prior network takes as input (1) occupancy map and (2) window of odometry measurements, and outputs a map of likelihood scores. Scores represent map locations that are likely given input odometry measurements. Scores used to weigh samples of a particle filter for tracking agent's trajectory.}
           \label{fig:overall_system_diagram}
\end{figure}

\noindent\textbf{Model Architecture:}
Our learnable Map Prior network is shown in Fig. \ref{fig:overall_system_diagram}. The network aims to generate the likelihood score that a given series of odometry measurements ends at a given location $\mathbf{x}$ in the map.
The likelihood score can be used directly as the particle filter weight for a particle at $\mathbf{x}$. We set up this network with two branches to separate map and odometry processing, whose deep encodings are later combined. This setup is useful because maps are generally static, so at runtime a specific indoor map only needs to be processed by the network once to generate its encoding. The stored map encoding can later be combined with the constantly changing odometry encoding, reducing computational burden. The inputs to our method are a $2$D indoor occupancy map $\mathnormal{M}$ and a short history of the agent's most recent odometry $\mathnormal{O}$ from time $t-n$ to $t$.
$n$ is selected experimentally to be $5$ seconds for human walking and $20$ seconds for slower robot driving. 
Our map network $f(\mathnormal{M}, \Theta_f)$ is a small version of U-Net \cite{ronneberger_u-net_2015, padhy_shreyaspadhyunet-zoo_2021} parameterized by weights $\Theta_f$ which has been modified to output a ``Deep Map Tensor'' of the same spatial dimensions as the input map, but with more channels. 
The number of channels $c=32$ is chosen experimentally, balancing speed of computation with sufficient space to store relevant environmental details. 
The odometry network $g(\mathnormal{O}, \Theta_g)$ is a two-layer Long-Short-Term Memory (LSTM) network parameterized by weights $\Theta_g$ operating on the time dimension of the odometry history. 
The LSTM hidden state is of size $c$, and the last hidden state output is used as the ``Deep Trajectory Vector''. 
A dot product is taken between the Deep Map Tensor and the Deep Trajectory Vector at each location $\mathbf{x}$, giving the likelihood score of $\mathbf{x}$ given $\mathnormal{O}$ and $\mathnormal{M}$:
\begin{align}
    \mathnormal{S}(\mathbf{x} | \mathnormal{M}, \mathnormal{O}) = f(\mathnormal{M}, \Theta_f)_{\mathbf{x}} \cdot g(\mathnormal{O}, \Theta_g)
\end{align}

\noindent\textbf{Model Training:}
\label{sec:model_training}
To train our model, we window ground truth walking trajectory data into windows of $n$ timestamps, and subtract the initial position in each window to make this data look like odometry data.
We additionally augment the training data and add a multiplicative velocity bias drawn from $\mathcal{N}(1, 0.5)$ pixel/second and a random noise drawn from $\mathcal{N}(0,0.25)$ pixels to each input sample to better mimic noisy odometry data. 
These windows are used as odometry branch training input.
The map branch is given small segments of the relevant map at train time to avoid overfitting to the most commonly feasible areas of the full map.

\begin{figure}[ht]
    \centering
    \begin{subfigure}{0.6\linewidth}
        \centering
      \includegraphics[width=\textwidth]{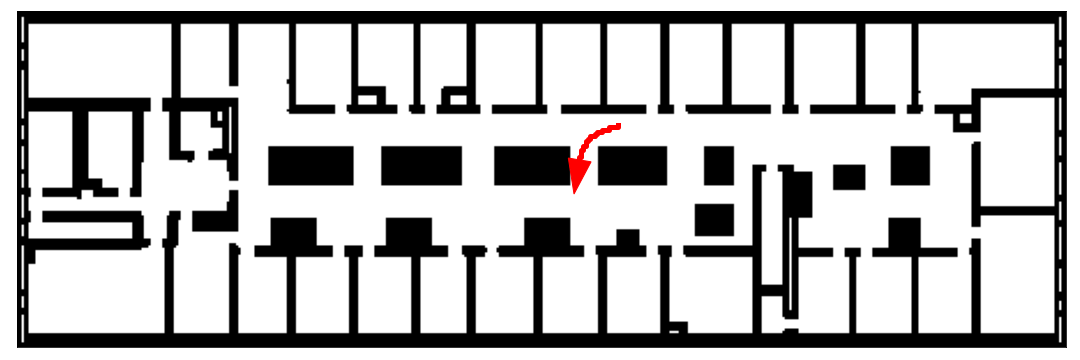}  
      \caption{Example input ground truth trajectory window (in red).}
      \label{fig:map_conv_traj_map}
    \end{subfigure}
    \hspace{1em}
    \begin{subfigure}{0.18\linewidth}
      \centering
      \includegraphics[width=\textwidth]{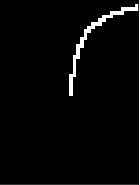}
      \caption{Cross-correlation kernel.}
      \label{fig:map_conv_kernel}
    \end{subfigure}
    \begin{subfigure}{0.6\linewidth}

        \centering
      \includegraphics[width=\textwidth]{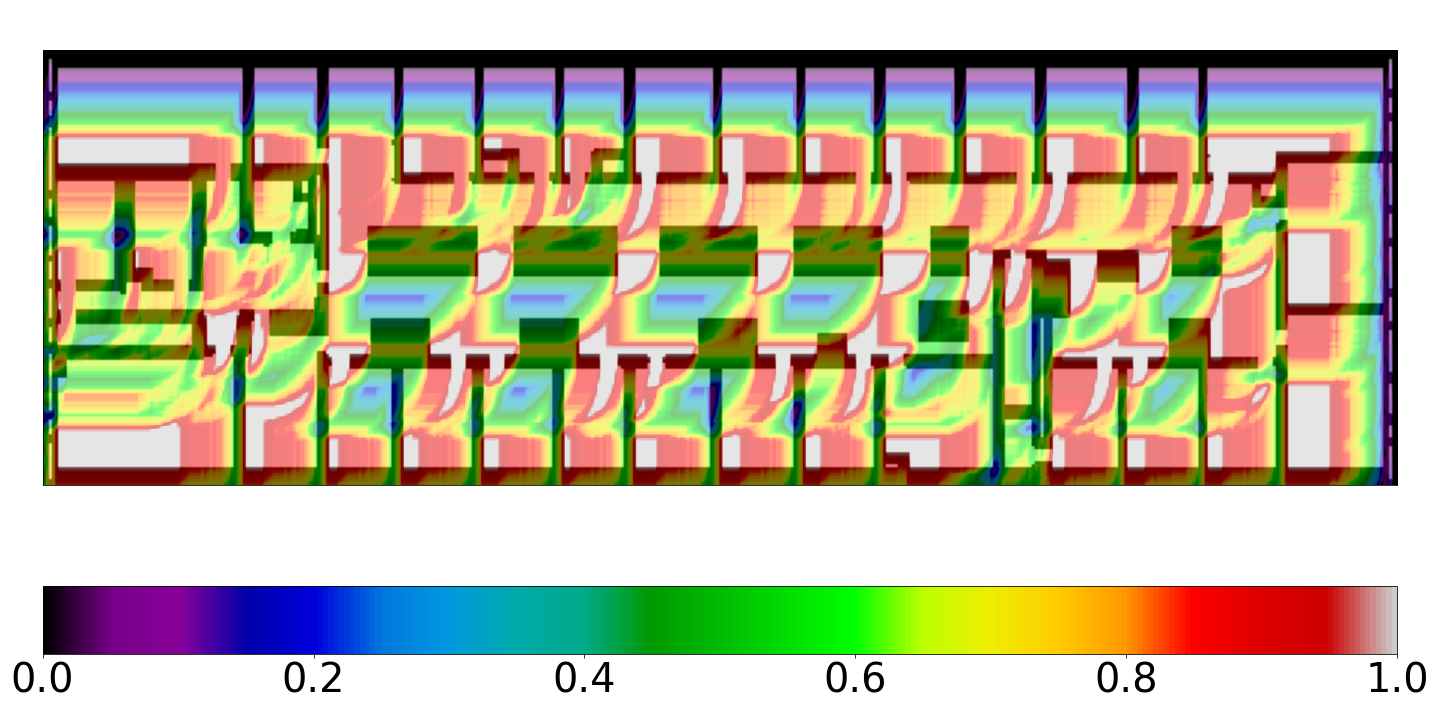}
      \caption{Resulting training target obtained by cross-correlating the kernel with the input map. Values near 1 are feasible trajectory end locations.}
      \label{fig:map_conv_result}
    \end{subfigure}
    
\caption{Ground truth generation process for training our map prior network.}
\label{fig:map_conv}
\end{figure}

We create a target $\mathnormal{T}$ (Fig. \ref{fig:map_conv_result}) defining which areas of the map are feasible end-locations for a given ground-truth trajectory window. We train our network to output $\mathnormal{T}$, even when the odometry is given as input is noisy or drifting. 
A 2D kernel like in Fig. \ref{fig:map_conv_kernel} is created from the ground truth trajectory segment and is cross-correlated with the map to form $\bar{\mathnormal{T}}$ whose values depend on the input trajectory's overlap with obstacles. $\mathnormal{T} = 10^{-6}*e^{14*\bar{\mathnormal{T}}}$ is computed to heavily upweight areas where the trajectory window passes only through free space, \textit{i.e.} is physically feasible. We train using Mean Squared Error (MSE) loss: $\mathcal{L} = \text{MSE}(\mathnormal{T}, \mathcal{S}(\boldsymbol{x}|\mathnormal{M}, \mathnormal{O}))$. In training our method we noticed that smaller feasible regions of $\mathnormal{T}$ were not being predicted well. We weight feasible regions by the inverse of their pixel area in our loss to improve small area prediction.

\noindent\textbf{Particle Filter:}
We utilize a particle filter to combine the learned map prior with odometry measurements, as particle filters have relatively low computational burden but are able to handle complex distributions of estimated location. A particle filter maintains $p$ estimates of the user's state (location) as particles. At each timestep, each particle has its stored state updated according to the motion model:
\begin{align}
    \mathbf{x}_{t+1} = \mathbf{x}_t + \Delta \mathbf{x} + \eta_{\text{motion}}
\end{align}
where $\mathbf{x}_t$ is the user's state $(x,y)$ at timestep $t$, and $\Delta \mathbf{x}$ is the odometry measurement between times $t$ and $t+1$. $2$D Gaussian noise $\eta_{\text{motion}} \sim \mathcal{N}(\mu, \Sigma)$ is added to each particle's motion to capture odometry uncertainty.

For our wheeled robot experiments, due to the significantly higher heading error present in wheel encoder odometry, we extend the state to also include robot heading $\theta$:
\begin{align}
    \nonumber s &= \Vert(\Delta x, \Delta y)\Vert_2\\
    \nonumber \mathbf{x}_{t+1} &= \mathbf{x}_{t} + s\begin{bmatrix}
    \cos(\theta)\\
    \nonumber \sin(\theta)\end{bmatrix}+ \eta_{\text{motion}(0,1)}\\
    \theta_{t+1} &= \theta_t + \Delta \theta + \eta_{\text{motion}(2)} 
\end{align}
The subscripts of $\eta$ indicate the respective components of the (now) $3$D noise vector, and $\Delta$ terms are odometry measurements. At test time for robot localization, position odometry is rotated to match the estimated heading before being passed to the map prior network to fix heading drift.

Particle weights are computed using the value of the location likelihood score heatmap at the particle's location from the last $n$ seconds of odometry. 
Low variance resampling of the particles is used due to its stability and consistent particle space coverage \cite{thrun_probabilistic_2005}. 
Resampling copies particles from the original cloud with probability proportional to their weight. 
The particle closest to the median (as opposed to the mean) of the new particle cloud is used as the agent's estimated location.
This guarantees the estimated location is not inside an obstacle even with a multi-modal particle distribution. 

We also implement a re-initialization method for situations where the particle filter has diverged from a reasonable estimate. 
We reinitialize by sampling particles randomly within a radius $r_\text{reinit}$ of the last estimated location when more than $s_\text{reinit}$ of particles have passed through an obstacle. 
We experimentally set $r_\text{reinit} = 5$ meters and $s_\text{reinit} = 90$\%.

\section{Datasets}
We show the performance of our method on three datasets with varied sources of odometry.

\noindent\textbf{IDOL Dataset:}
The IDOL dataset \cite{sun_idol_2021} is an inertial odometry dataset for indoor pedestrian localization. 
It consists of $20$ hours of pedestrian trajectories collected by users carrying a smartphone in $3$ buildings. 
Smartphone IMU data and ground truth device pose is available at $100$ Hz. Odometry is generated from IMU data by the IDOL method proposed in the same work. 
$2$D occupancy maps are generated from architectural plans available for the $3$ buildings.

\begin{figure}[ht]
    \centering
    \subfloat[IDOL dataset\label{fig:idol_rig}]{
         \includegraphics[width=0.45\linewidth]{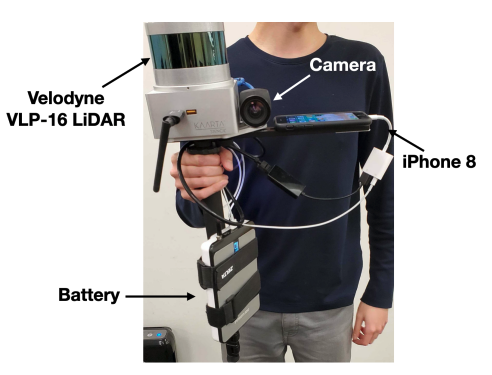}
     }
     \subfloat[BLE + IMU dataset\label{fig:zed_rig}]{
         \includegraphics[width=0.35\linewidth]{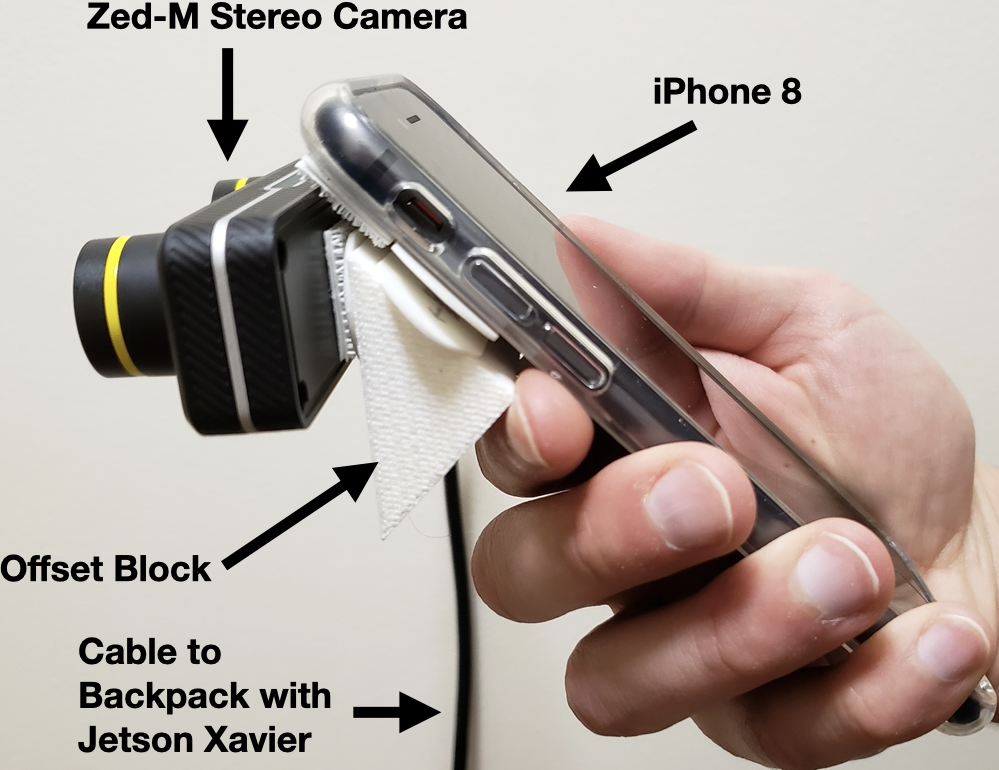}
     }
     \hspace{1em}
     \subfloat[TurtleBot dataset\label{fig:tb_rig}]{
         \includegraphics[width=0.35\linewidth]{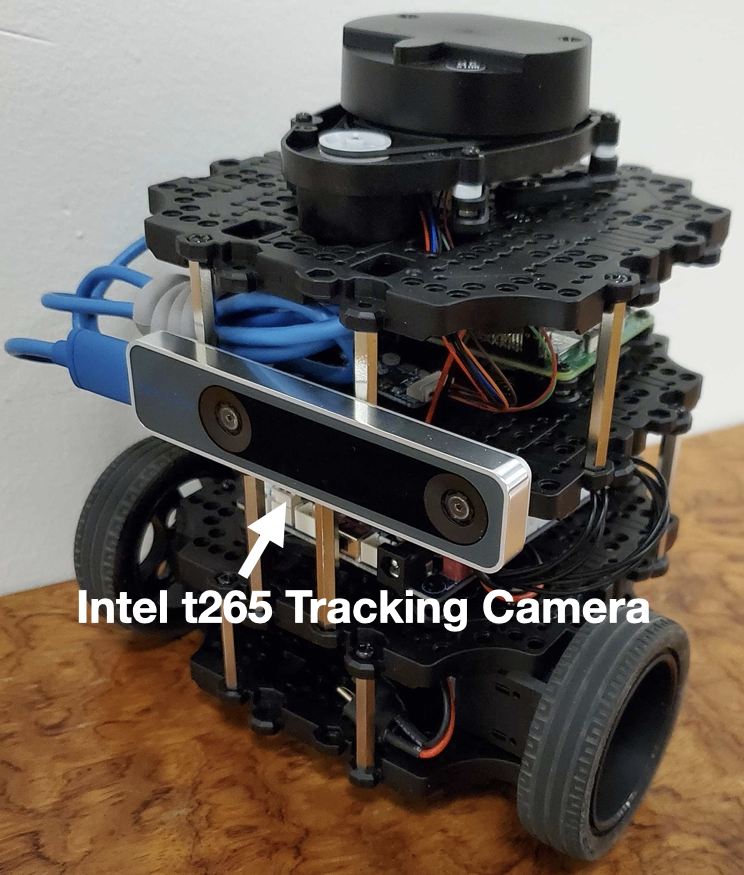}
     }
     \caption{Data collection setups used in our experiments.}
     \label{fig:rigs}
 \end{figure}

\noindent\textbf{BLE + IMU Dataset:}
We collect a smaller dataset similar to the IDOL dataset but with the addition of Bluetooth low energy (BLE) beacon measurements, to compare against bluetooth-based localization.
Two floors of a building like those used in the IDOL dataset are instrumented with BLE beacons emitting a unique identifier broadcast at $1$ Hz, which are recorded along with the broadcast signal strength and IMU data by a smartphone at $60$ Hz. 
Ground truth device pose and orientation are generated with a Stereolabs Zed Mini stereo camera configured as in Fig. \ref{fig:zed_rig}. 

\noindent\textbf{TurtleBot Dataset:}
A ROBOTIS TurtleBot 3 (Fig. \ref{fig:tb_rig}) was driven in $2$ maps to collect $8$ trajectories each of $10$ minutes duration to evaluate generalization to wheel encoder odometry. 
Wheel odometry was generated using the onboard wheel encoder and IMU data, and ground truth pose was recorded from an attached Intel T$265$ tracking camera. The robot state $(x,y,\theta)$ follows a standard differential drive model:
\begin{align}
    \nonumber \Delta s &= \pi r (n_{L} + n_{R}) \\
    \nonumber x_{t+1} &= x_t + \Delta s \cos(\Delta \theta + \nonumber \theta_{t})\\
    \nonumber y_{t+1} &= y_{t} + \Delta s \sin(\Delta \theta + \theta_{t})\\
    \theta_{t+1} &= \theta_t + \Delta \theta
\end{align}
where $r$ is the wheel radius, $n_{L}$ and $n_{R}$ are the number of revolutions recorded by the wheel encoders since the last time step, and $\Delta \theta$ is the relative orientation computed via a standard Madgwick filter \cite{madgwick_estimation_2011} from IMU data.

\section{Experiments}
\label{sec:result}

We evaluate our method in three experimental settings. 
(1) \textbf{IMU only} where we show the ability of our learned map prior to reduce drift in IMU-only odometry better than heuristic map priors. 
(2) \textbf{BLE + IMU} where we show that our IMU only method performs at a similar level to a BLE + IMU method that utilizes absolute positioning. 
(3) \textbf{Turtlebot} where we show the ability of our map prior network to generalize to unseen wheel encoder odometry.

\noindent\textbf{Metrics:} We evaluate localization performance using two main metrics: Absolute Trajectory Error (ATE)  and End Error (EE). 
ATE is the root mean square error between points in the estimated and ground truth trajectories. The error at timestamp $i$ is $e_i = \Vert\mathbf{x}_i - \hat{\mathbf{x}}_i\Vert_2$. ATE measures global trajectory consistency and tends to increase with time due to drift. EE is the distance between the final estimated and final ground truth positions in a trajectory, i.e. $e = \Vert\mathbf{x}_{T} - \hat{\mathbf{x}}_T\Vert_2$. EE measures the total accumulated drift in the sequence.

\noindent\textbf{Training/Testing:}
We implement the learnable map prior model in Pytorch Lightning \cite{falcon_pytorchlightningpytorch-lightning_2020} and train on a single Nvidia RTX $2080$ Ti, with the Adam optimizer \cite{kingma_adam_2017} and a learning rate of $0.01$. 
A batch size of $32$ is used and the network takes between $50$ and $100$ full passes through the data (epochs) to converge, depending on the map.

At test time the particle filter initializes with $1000$ normally distributed particles ($\sigma = 0.01$m) around the true starting location. 
The filter runs at $1$ Hz. 
For inertial odometry, the motion model uses white Gaussian noise with $\Sigma = 0.1\mathcal{I}_2$ m. $\Sigma = 0.01\mathcal{I}_3$ m is used for wheeled encoder odometry.
A benefit of our formulation is the decomposition of trajectory and map processing. 
Once the network is trained, the map only needs to be processed once to obtain a Deep Map Tensor. 
This can be reused for all trajectories in the same map. 
At test time, only the trajectory LSTM needs to process data.
An AMD Threadripper $1920$x CPU takes $6.7$ ms on average to compute the location likelihood score heatmap using a stored Deep Map Tensor. 
Our non-optimized particle filter runs at $4$x real-time speed, suggesting that our method would be feasible for smartphones or low-compute platforms.

\noindent\textbf{IMU Only Experiments:}
In our first experiment, we compare our method against four other localization methods to show the utility of our learned map prior. (1) Pedestrian Dead Reckoning (\textbf{PDR}) is implemented similarly to the baseline used in \cite{yan_ronin_2019}. Step size is assumed to be $0.67$ m, and the iOS internal estimate of device heading is used. (2) \textbf{IDOL} \cite{sun_idol_2021} is used to represent deep inertial odometry performance. We also compare against two approaches that use heuristic map priors. (3) One is a particle filter (\textbf{Heuristic PF}) implementation based on \cite{rechy_rormero_map-aware_2018}, where the prior is computed as the map cross correlation with a kernel based on noisy odometry. (4) The other is a linear-chain \textbf{CRF} implementation based on \cite{xiao_lightweight_2014} that uses a hand-designed graph of possible user locations and transitions, and feature functions that capture the probability of transitions based on the observations. We use a unary feature function based on the absolute position and a pair-wise feature function based on relative odometry displacement. The feature function weights and graph edge length are chosen using grid search to achieve a good trade-off between accuracy and runtime.

\begin{figure}[h]
    \centering
    \includegraphics[width=\linewidth]{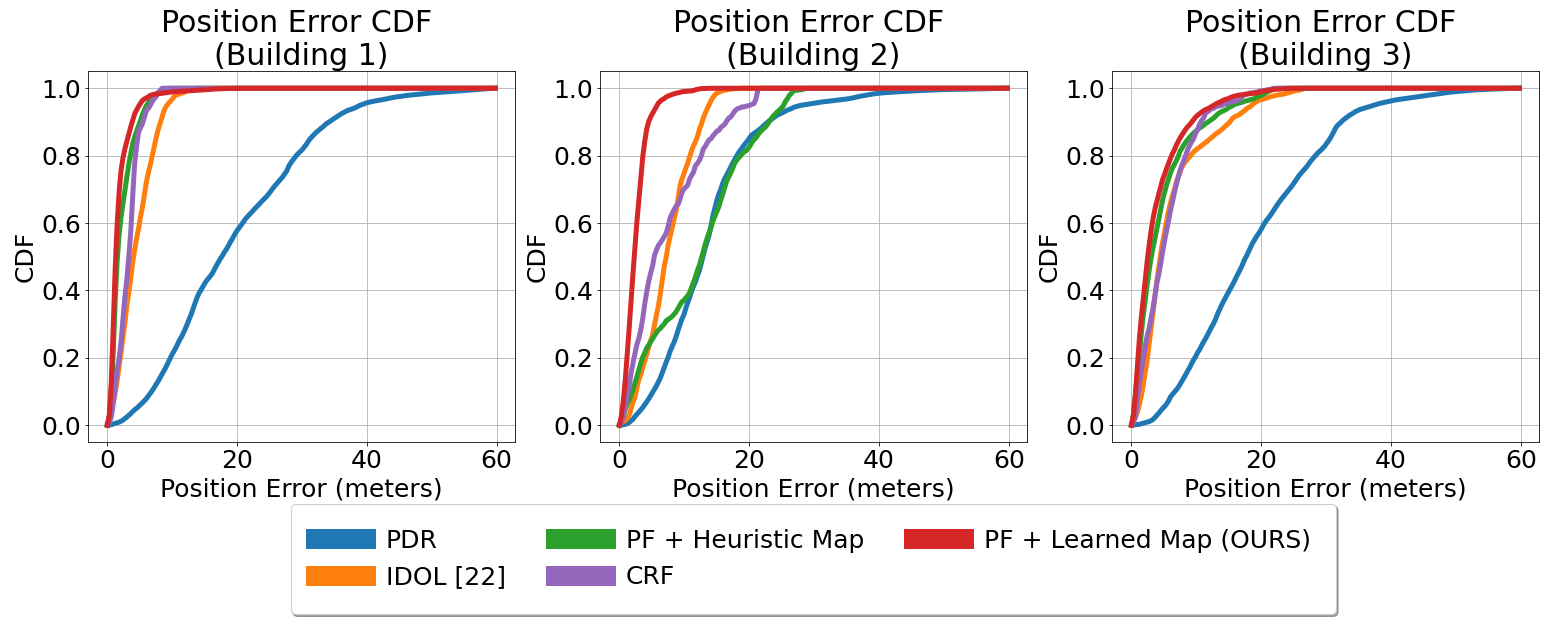}
    \caption{Cumulative distribution function ATE error of our method and baselines from IDOL dataset experiments.}
    \label{fig:IDOL_CDFs}
\end{figure}

Tab. \ref{tab:position_error_idol} shows the results of our localization performance on the IDOL dataset. Fig. \ref{fig:IDOL_CDFs} shows the error distributions. Our method significantly outperforms pure odometry and heuristic map-based methods across buildings in both ATE and EE. Fig. \ref{fig:IDOL_trajs} shows the qualitative improvement in localization accuracy using our method.
On average our method is able to achieve an error of less than $4$ meters $80\%$ of the time.

Fig. \ref{fig:IDOL_trajs} shows that our method estimates more accurate locations and does not exhibit the large errors observed in heuristic methods. 
Notice that the heuristic PF often generates trajectories that penetrate walls similar to the odometry-only trajectory (Building 2).
Building 2 has a long hallway which the odometry often drifts out of as users walk down it. 
Due to only small overlap of the trajectory with the walls, the heuristic prior weights the hallway-adjacent rooms similarly to the hallway. 
This leads to drift out of the hallway into these rooms.
While the CRF trajectory does not pass through walls due to map constraints, it can erroneously enter rooms and ultimately become stuck. 
The CRF also outputs sharp unnatural trajectories because of the discretization of the map. 
Reducing the step resolution to generate a finer graph causes the inference to quickly becomes infeasible for real-time operation as the computation time grows quadratically with graph size. 
Building 3 is substantially larger than buildings 1 and 2 and has narrow hallways. These factors lead to significant odometry drift which our method is able to handle consistently, except when the trajectory deviates too far away from the map which results in a slightly inferior EE accuracy.

\begin{table}[ht]
    \centering
    \begin{tabular}{lrrrrrrrrrrrr}
        \toprule
        \multirow{2}{*}{\textbf{Model}} & \multicolumn{2}{c}{Bldg 1} & \multicolumn{2}{c}{Bldg 2} & \multicolumn{2}{c}{Bldg 3}\\
        \cmidrule(lr){2-3}
        \cmidrule(lr){4-5}
        \cmidrule(lr){6-7}
        & \textbf{ATE} & \textbf{EE} & \textbf{ATE} & \textbf{EE} & \textbf{ATE} & \textbf{EE}\\
        \midrule
        PDR & 24.28 & 24.96 & 12.66 & 13.80 & 21.86 & 22.18 \\
        IDOL & 5.65 & 8.51	&6.62 &10.61 &8.33 & 9.71 \\
        CRF & 4.66 & 6.58 & 7.56 & 10.54 & 8.82 & \textbf{7.38}\\
        Heuristic PF &3.11  &	3.42 & 11.37 &15.36	&6.71& 9.67	 \\
        Ours & \textbf{2.87} & \textbf{1.60}& \textbf{2.51} & \textbf{6.08}& \textbf{5.66} & 9.99	 \\
        \bottomrule
    \end{tabular}
    \caption{Quantitative results of our approach compared to baselines on the IDOL\cite{sun_idol_2021} dataset. We report mean trajectory errors (in meters) over all sequences per building.}
    \label{tab:position_error_idol}
\end{table}

\begin{figure}[h]
    \centering
    \vspace{-12px}
    \includegraphics[width=\linewidth]{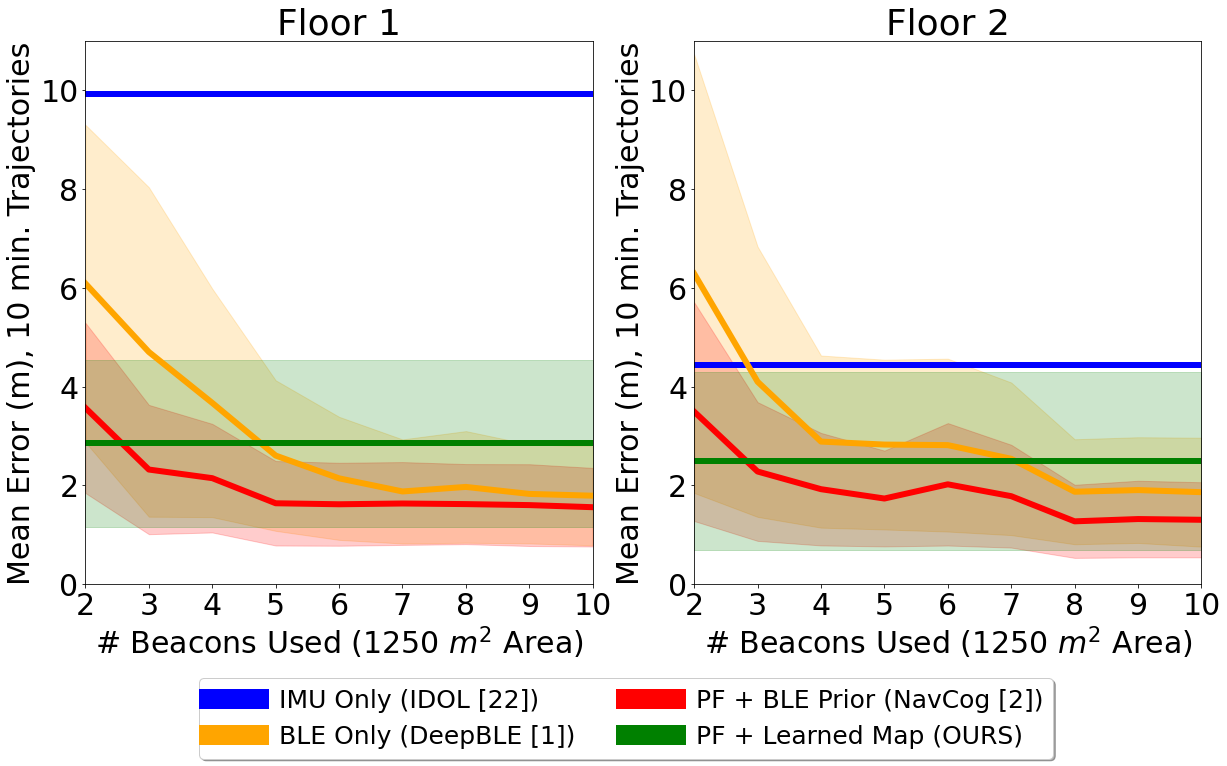}
    
    \caption{Comparison of mean trajectory errors for varying bluetooth beacon densities for BLE + IMU experiments. Std. devs. shown by shaded regions. IDOL and our method do not use beacons so are independent of beacon density. Our method achieves accuracy similar to bluetooth-based absolute positioning without instrumentation of the environment.}
    \label{fig:nsh_ble_imu_comp}
\end{figure}

\noindent\textbf{BLE + IMU Experiments:}
We now compare our method using relative position measurements and a map to an absolute positioning method which directly estimates a user position in a global reference frame, \textit{i.e.} BLE localization. BLE localization is implemented in two ways: (1) pure deep BLE localization (\textbf{BLE Only}) as in \cite{agarwal_deepble_2021}, and (2) using the deep BLE estimates in the particle filter observation model instead of a map (\textbf{PF + BLE Prior}) similar to NavCog \cite{ahmetovic_navcog_2016}.

Fig. \ref{fig:nsh_ble_imu_comp} shows the results of using various numbers of beacons in a beacon localization system compared to IDOL and our method. 
BLE-only uses the raw, $1$ Hz BLE estimates of location. 
PF + BLE Prior is able to achieve better performance by smoothing BLE estimates using inertial odometry (\textit{i.e.} IDOL). 
Inertial odometry alone generally does not outperform BLE localization, but with the addition of map information through our learned prior we outperform BLE-only localization with fewer than $5$-$7$ beacons.
Overall, we achieve ATE within $\sim 1$ meter of both BLE methods using $8$-$10$ beacons.
Beacon densities of $8$-$10$ have the best performance since many beacons can be used to precisely triangulate the user.
These experiments are performed in a $\sim1250$ m\textsuperscript{2} area, which is fairly small relative to $10$ beacons.
Similar performance in larger areas would require hundreds of beacons which is not scalable (for both installation and maintenance).
Our method is slightly less accurate but does not require instrumentation of the environment.

\begin{figure}[h]
    \centering
    \includegraphics[width=\linewidth]{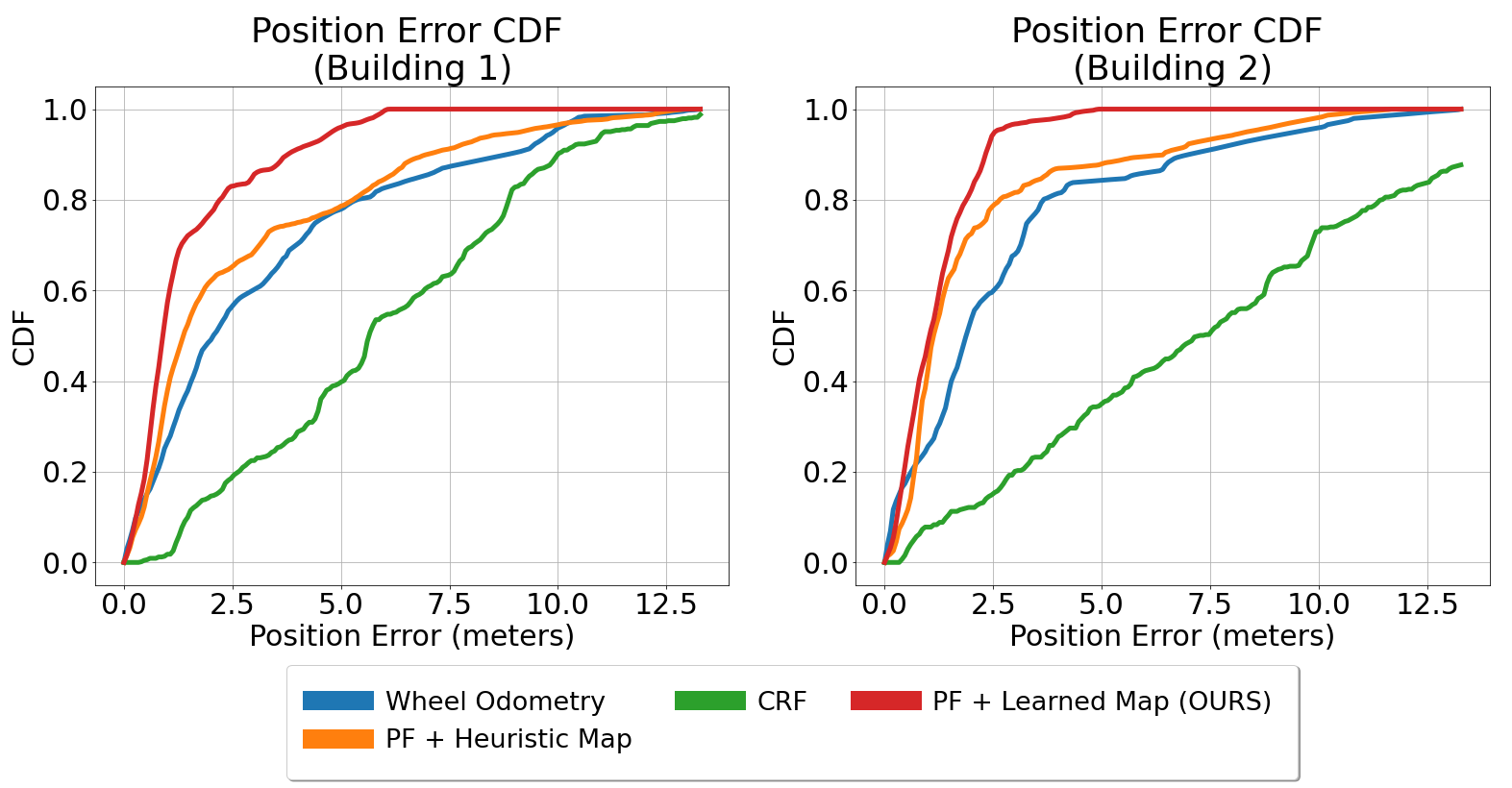}
    \caption{Cumulative distribution function of ATE of our method and baselines from the Turtlebot experiments. Our method achieves best performance even without retraining on wheel odometry.}
    \label{fig:ROBOT_CDFs}
\end{figure}

\begin{table}[ht]
    \centering
    \begin{tabular}{lrrrr}
        \toprule
        \multirow{2}{*}{\textbf{Model}} & \multicolumn{2}{c}{Bldg 1} & \multicolumn{2}{c}{Bldg 2} \\
        \cmidrule(lr){2-3}
        \cmidrule(lr){4-5}

        & \textbf{ATE}  & \textbf{EE} & \textbf{ATE}  & \textbf{EE} \\
        \midrule
        Odometry &  4.15&	7.84 &	3.93 &	3.99  \\
        \midrule
        CRF & 3.33 & 6.31 & 3.91 & 4.36\\
        \midrule
        Heuristic PF &3.69 &6.25 &3.20 &2.89  \\ 
        \midrule
        Ours &  \textbf{1.88}&	 \textbf{1.69}&	\textbf{1.48} & \textbf{1.48}	  \\
        \bottomrule
    \end{tabular}
    \caption{Quantitative results of our approach in comparison with baselines on the Turtlebot dataset. We report the mean trajectory errors (in meters) over all the sequences in a building.}
    \label{tab:position_error_robot}
\end{table}
\noindent\textbf{Turtlebot Experiment:}
Tab. \ref{tab:position_error_robot} reports results of our approach in tracking Turtlebot trajectories using wheel encoder odometry. Fig. \ref{fig:ROBOT_CDFs} shows error distributions. Our learned prior in a particle filter halves localization error across buildings, which the heuristic methods are unable to handle. On average our method is able to achieve an error of $2.8$ meters $90\%$ of the time. This performance is achieved without retraining the network which has only ever seen inertial odometry.

\begin{figure}[h]
    \centering
    \includegraphics[width=\linewidth]{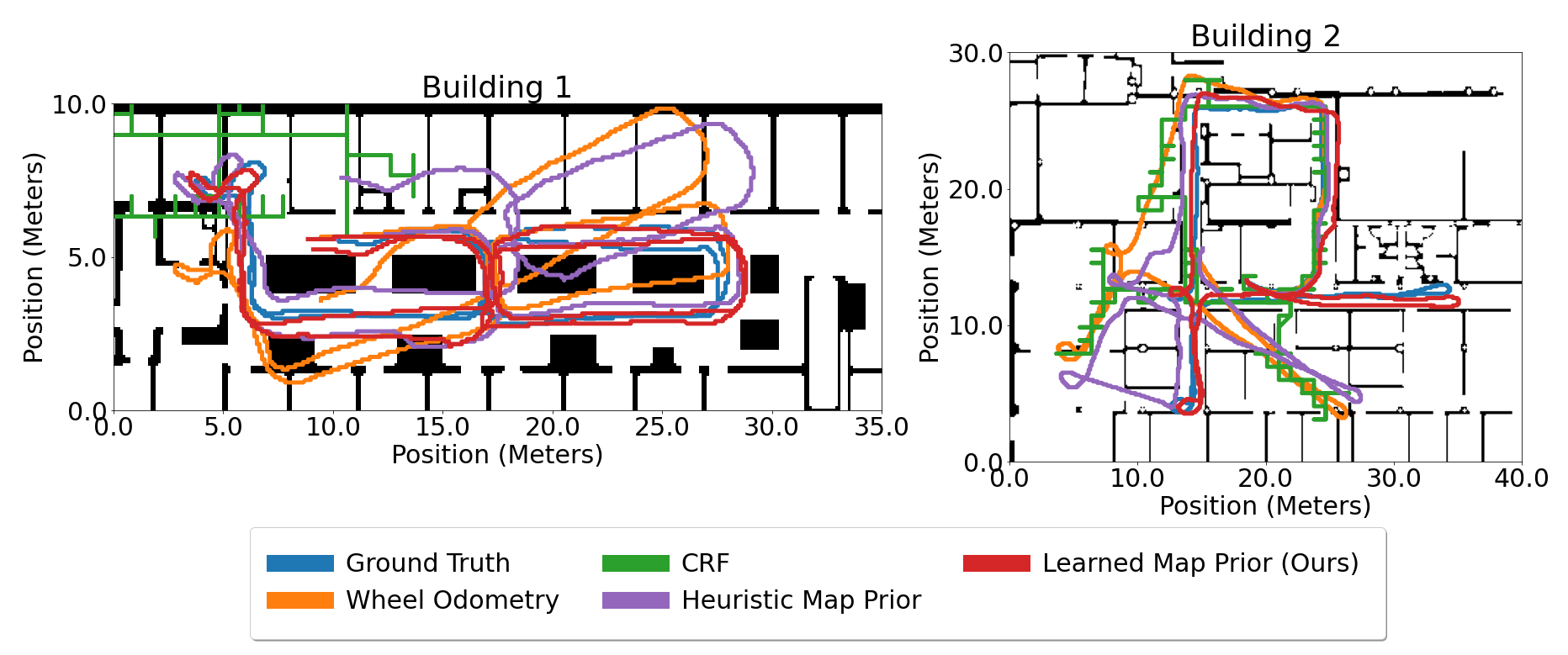}
    \caption{Qualitative results showing the trajectories of the Turtlebot estimated by our method and other baselines.}
    \label{fig:ROBOT_trajs}
\end{figure}

The trajectories in Fig. \ref{fig:ROBOT_trajs} show our prior's ability to accurately weigh particles.
Odometry heading error often accumulates as the robot makes turns, but our learned prior absorbs some of this error and places weight in appropriate locations based on the general curving nature of the robot's trajectory.
The heuristic PF is unable to accurately weight regions when heading error occurs and so does not correct the error, leading to significant drift. The curvy nature of the trajectory also poses problems for the CRF as it is unable to capture these smooth transitions on the graph and instead estimates sharp lateral transitions which accumulate more error.

\begin{figure}[h]
    \centering
    
    \begin{subfigure}{0.64\linewidth}
        \includegraphics[width=\linewidth]{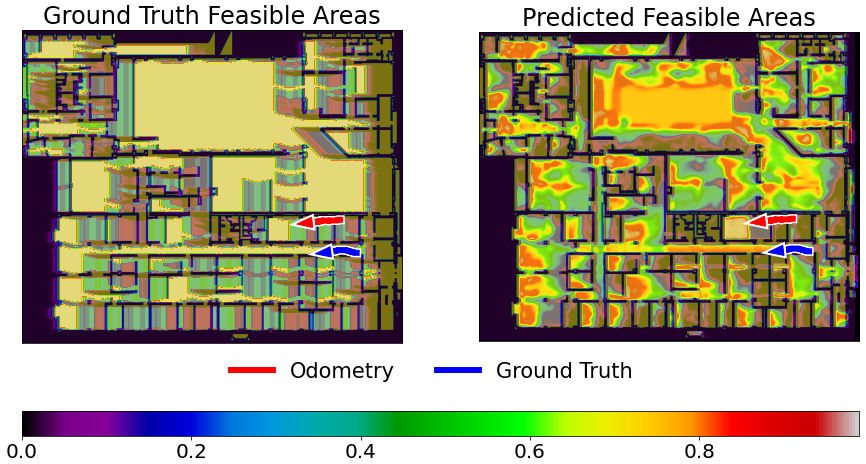}
        \caption{Example map prior prediction (based on red odometry trajectory) next to the ground truth feasible locations (generated from the ground truth trajectory in blue).}
        \label{fig:map_prior_examples}   
    \end{subfigure}
    \hfill
    \begin{subfigure}{0.31\linewidth}
         \centering
        \includegraphics[width=\linewidth]{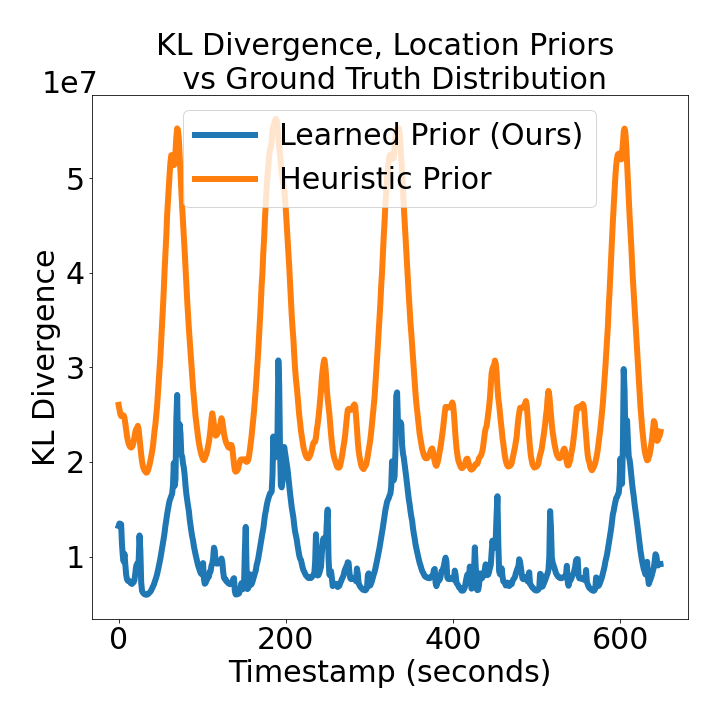}
        \caption{KL divergence between ground truth user location distribution and the tested map priors.}
        \label{fig:kl_div}       
    \end{subfigure}
    
    \caption{Map prior visualization.}
    \label{fig:map_prior_analysis}

\end{figure}

\noindent\textbf{Learned Map Prior Visualization:}
Fig. \ref{fig:map_prior_examples} shows a visualization of the map prior network output $\mathcal{S}(\mathbf{x} | \mathnormal{M}, \mathnormal{O})$ for an input trajectory window. Spaces where the trajectory window input is feasible are generally given higher weight than extremely unlikely areas. 

We also compare the KL divergence between a ``ground truth'' location distribution $G$ and our predicted prior distribution, and between $G$ and the heuristic prior distribution, as shown in Fig. \ref{fig:kl_div}.
This indicates how well each prior places weight in the ground truth user location and avoids placing weight in unlikely locations.
$G$ is defined as a Gaussian ($\sigma = 1$m) around the ground truth user location. 
Our learned prior shows consistently lower divergence from the ground truth than the heuristic prior by a factor of approximately 3.

\section{Conclusion}
\label{sec:conclusion}

In this work, we present a data-driven approach to combine occupancy map information with odometry measurements for indoor localization. 
Existing work uses hand-defined methods to incorporate map information. 
These methods make assumptions which generalize poorly or use noise-sensitive heuristics.
Our method provides a more robust prior on indoor user location using a two-branch deep network utilizing separate learned spatial map and temporal odometry embeddings which are combined to determine feasible user locations in the map.
Our prior, when used as a sensor model in a particle filter, is able to achieve $49\%$ accuracy improvement over odometry for pedestrian localization.
Compared to a BLE beacon-based localization system, our method exceeds BLE accuracy in low beacon-density spaces and approaches similar performance to BLE methods in high beacon-density spaces without requiring any devices installed in the environment.
Our prior is also versatile, halving error in wheel encoder odometry-based localization for a robot without retraining. 
A key avenue of future work will be addressing generalization of our map network to unseen maps through data collection in varied maps and synthetic data-generation techniques.

\bibliography{references}
\footnotesize{
\bibliographystyle{IEEEtran}
}
\end{document}